\documentclass{article}
\usepackage{preprint,times}
\preprintfinalcopy

\usepackage{hyperref}
\usepackage{url}
\usepackage{graphicx}
\usepackage{wrapfig}
\usepackage{booktabs}
\usepackage{colortbl}
\usepackage{amsmath,amssymb}
\usepackage{multirow}
\usepackage{placeins}
\usepackage{capt-of}
\usepackage{pifont}
\usepackage{xcolor}
\usepackage{xspace}
\usepackage[most]{tcolorbox}
\definecolor{taskgreen}{HTML}{1F3A5F}
\newtcolorbox{taskbox}[1]{enhanced, colback=white, colframe=taskgreen, boxrule=0.6pt, arc=2pt,
 left=4pt, right=4pt, top=3pt, bottom=3pt, fonttitle=\bfseries\small, coltitle=white, colbacktitle=taskgreen,
 title={#1}, before skip=3pt, after skip=3pt, fontupper=\footnotesize}

\usepackage{amsmath,amsfonts,bm}

\def\eqref#1{equation~\ref{#1}}

\def\1{\bm{1}}

\DeclareMathAlphabet{\mathsfit}{\encodingdefault}{\sfdefault}{m}{sl}
\SetMathAlphabet{\mathsfit}{bold}{\encodingdefault}{\sfdefault}{bx}{n}

\graphicspath{{figures/}}

\newcommand{\method}{R\textsuperscript{2}-WAM\xspace}
\title{Data-Free Continual Learning for World Action Model with Rectified Generative Replay}

\author{\begin{minipage}{\textwidth}\centering\vspace{10pt}
Kaipeng Fang$^{1,2}$ \quad Hangxu Liu$^{2}$ \quad Chaojian Shi$^{3}$ \quad Yunsong Zhou$^{2\S}$ \\[2pt]
Hongzi Zhu$^{4}$ \quad Minyi Guo$^{4}$ \quad Jingkuan Song$^{5,6}$ \quad Lianli Gao$^{1\S}$ \\[6pt]
{\mdseries\small
$^{1}$University of Electronic Science and Technology of China \quad
$^{2}$Shanghai AI Lab \quad
$^{3}$Fudan University \\[1pt]
$^{4}$Shanghai Jiao Tong University \quad
$^{5}$Tongji University \quad
$^{6}$Shanghai Innovation Institute}
\end{minipage}}

\everydisplay{\abovedisplayskip=4pt plus 1pt minus 2pt \belowdisplayskip=4pt plus 1pt minus 2pt \abovedisplayshortskip=2pt \belowdisplayshortskip=2pt}
\begin{document}

\maketitle
\lhead{} \renewcommand{\headrulewidth}{0pt}
{\renewcommand{\thefootnote}{\S}\footnotetext{Equal advising.}}

\begin{abstract}
World action models (WAMs) built on pretrained video generation models adapt quickly to new robotic tasks, but they forget previously learned tasks under sequential fine-tuning.
Experience replay (ER) mitigates such forgetting by rehearsing data of previous tasks, yet this data is often unavailable in robotics, where a policy is deployed as weights only and its training data is not shared along with it.
In this work, we ask whether a WAM can achieve data-free continual learning by regenerating its own past.
Our key observation is that a WAM can indeed autoregressively regenerate complete rollouts of previously learned tasks and replay them to mitigate forgetting.
However, replaying such raw rollouts still lags far behind replaying real data, which we attribute to two issues.
First, the WAM is unable to recognize task completion and keeps generating purposeless motion after the task is done.
Second, under open-loop generation, the generated actions gradually drift away from the generated observations.
To address both issues, we present rectified generative replay (\method), which truncates generated rollouts at task completion with an agentic evaluator and filters out rollouts whose actions are physically inconsistent with the observations.
On the four LIBERO suites and five real-world tasks, \method reduces forgetting by over 90\% relative to sequential fine-tuning and nearly matches ER without accessing any data from previous tasks.
Our analysis further reveals that forgetting in WAMs concentrates in inverse dynamics modeling: a forgotten WAM still produces visually plausible observations of previous tasks while its generated actions fail.
Project page: \url{https://kaipengfang.github.io/r2-wam/}

\end{abstract}

\section{Introduction}
\label{sec:intro}
\begin{wrapfigure}[24]{r}{0.52\textwidth}
\vspace{-0.3\baselineskip}
\centering
\includegraphics[width=\linewidth]{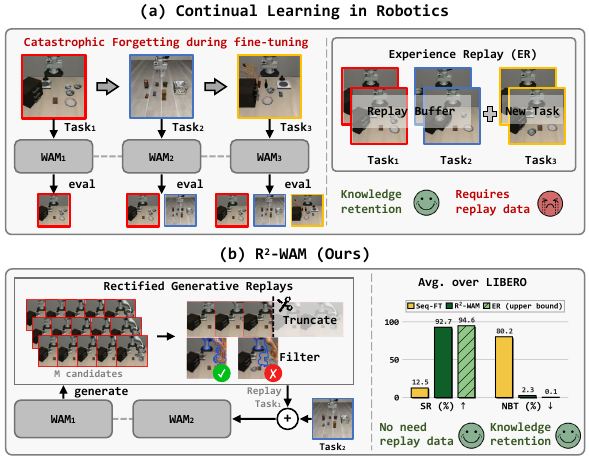}
\vspace{-1.0\baselineskip}
\caption{\textbf{(a)} In continual learning, a WAM is sequentially fine-tuned on new tasks and evaluated on all tasks learned so far. Experience replay (ER) mitigates forgetting of earlier tasks but requires access to their data. \textbf{(b)} \method exploits the autoregressive generation of WAMs to achieve data-free continual learning, and its rectified generative replay attains performance comparable to ER.}
\label{fig:teaser}
\end{wrapfigure}

Robots deployed in the real world keep encountering new tasks, so a practical policy has to learn them continually while retaining the tasks it has already learned, a capability known as continual learning~\citep{liu2023libero}. Among robot policies, world action models (WAMs) have emerged as a promising paradigm, as they leverage powerful pretrained video generation models~\citep{blattmann2023stablevideo,wan2025,nvidia2025cosmospredict2} to jointly predict future observations and robot actions~\citep{lingbotva2026,cosmospolicy2026,dreamzero2026}. We find that sequentially fine-tuned WAMs adapt well to new tasks yet collapse on previously learned ones, as shown in Fig.~\ref{fig:teaser}(a), i.e., they suffer from catastrophic forgetting.

To mitigate catastrophic forgetting, continual learning methods either constrain parameter updates or perform experience replay (ER). The former protects prior knowledge by regularizing~\citep{kirkpatrick2017ewc} or isolating~\citep{rusu2016pnn} parameters, but regularization retains little for robotic policies~\citep{liu2023libero}, and isolation needs a task identifier at test time. ER instead maintains a buffer of demonstrations from previous tasks and mixes them with new-task data during fine-tuning~\citep{rolnick2019er}, as shown on the right of Fig.~\ref{fig:teaser}(a), and remains among the most effective and robust choices for robotic policies~\citep{liu2023libero,liu2026vlaresist}. However, ER assumes that demonstrations accumulate along with the tasks, whereas a policy in practice often arrives as weights alone, and the demonstrations that trained it are no longer at hand. A policy should therefore learn continually in a \emph{data-free} manner, without access to any data from previous tasks. WAMs offer a potential way to achieve this, since they model how the world evolves and can autoregressively predict future observations and actions, rolling out complete trajectories of previously learned tasks for replay. This raises a natural question: \emph{Can a WAM leverage its own generative capability to achieve data-free continual learning as effectively as ER?}

In this work, we answer this question with rectified generative replay (\method), which regenerates rollouts of previous tasks with the WAM itself and rectifies them before replay, as shown in Fig.~\ref{fig:teaser}(b). Building on LingBot-VA~\citep{lingbotva2026}, we find that a WAM can indeed regenerate complete rollouts of previously learned tasks from their instructions alone, and replaying them already mitigates forgetting. However, replaying such raw rollouts still lags far behind ER, which we attribute to two issues. \textbf{First}, a WAM is unable to recognize task completion and receives no external stop signal, so generation runs for a fixed number of steps and the rollout keeps moving after the task is done, producing purposeless post-completion motion. \textbf{Second}, open-loop generation feeds the WAM's own predictions back without physical grounding, so errors accumulate and the generated actions gradually drift away from the generated observations. \method addresses both issues with completion-aware adaptive truncation and flow-based physical-consistency filtering. The former removes post-completion motion with an agentic evaluator that detects task completion and adaptively truncates each rollout. The latter checks whether the generated actions agree with the generated observations by comparing two optical flows of the robot: one simulated by executing the generated actions in a robot-only simulator, which is possible since the robot itself is fully specified by its URDF, and one estimated from the generated observations. We rank the rollouts by the agreement between the two flows and keep the most consistent ones for replay. The rectified rollouts are then mixed with new-task data for fine-tuning.

We evaluate \method on the four LIBERO suites~\citep{liu2023libero} and a real-world suite of five tasks. Under the data-free continual learning setting, \method reduces forgetting by over 90\% relative to sequential fine-tuning on both LIBERO and the real world, and recovers almost all of the retention of ER without any data from previous tasks. Our ablations confirm that both truncation and filtering are necessary, and our analysis further reveals that forgetting in a WAM concentrates in inverse dynamics modeling, whereas its ability to predict observations is largely preserved. As a result, a forgotten WAM still produces visually plausible observations while its generated actions fail, which also points to the action side as the part to verify.

Our main contributions are summarized as follows:
\begin{itemize}
 \item We show that, without any real data of previous tasks, a WAM can mitigate catastrophic forgetting by replaying its own generated rollouts and, once rectified, nearly match ER.
 \item We propose \method, which rectifies the generated rollouts before replay with completion-aware adaptive truncation and flow-based physical-consistency filtering.
 \item In simulation and the real world, \method eliminates over 90\% of the forgetting of sequential fine-tuning and comes within about 2~pp of ER without any data from previous tasks, and find that forgetting in WAMs concentrates in inverse dynamics modeling.
\end{itemize}

\section{Related Work}
\label{sec:related}
\paragraph{Continual learning in robotics.}
Continual learning methods mitigate forgetting by regularizing parameter updates or outputs~\citep{kirkpatrick2017ewc,zenke2017si,li2017lwf}, isolating or adding task-specific parameters~\citep{rusu2016pnn,mallya2018packnet,hu2022lora,wang2022l2p}, or storing a subset of previous-task data and rehearsing it, known as experience replay~\citep{rolnick2019er,chaudhry2019agem,caccia2022erace,wang2024survey}.
In robot learning, experience replay remains a strong baseline from LIBERO to recent VLA policies~\citep{liu2023libero,wu2025stellar,liu2026vlaresist}, while other methods rely on task-specific adapters~\citep{liu2024tail,romer2026clare}, skill libraries or prompts~\citep{wan2024lotus,yao2025ppl,zheng2025imanip}, or representation distillation~\citep{roy2025m2distill}.
\method follows the replay paradigm but stores no data from previous tasks, regenerating the replay data from their instructions instead, a setting known as data-free continual learning in image classification~\citep{yin2020deepinversion,smith2021abd}.

\paragraph{Generative replay.}
Unlike experience replay, which must store data from previous tasks, generative replay rehearses samples drawn from a generative model~\citep{robins1995pseudo,shin2017dgr}, either a separate generator whose outputs are labeled by conditioning or by an old copy of the learner~\citep{wu2018mergan,gao2023ddgr,kim2024sddgr,gao2021cril,yue2024tdgr,pan2025lifelongwm}, or the learner itself prompted to produce its own replay data~\citep{vandeven2020brain,sun2020lamol,huang2024ssr,wang2025sepr}.
Closest to our work, REGEN~\citep{govind2026regen} and Dream Rehearsal~\citep{nijjer2026dream} replay generated rollouts for continual imitation and reinforcement learning.
Both start generation from real observations, and Dream Rehearsal further scores the rollouts with the RL reward.
In contrast, \method generates rollouts from the instruction alone and rectifies them without any reward.

\paragraph{World action models.}
Building on pretrained video generation models, world action models jointly predict future observations and robot actions~\citep{wu2024gr1,guo2024pad,li2025uva,uwm2025,liao2026genie,cosmospolicy2026,dreamzero2026,lingbotva2026}, and have shown strong performance and generalization in manipulation.
Beyond serving as a policy, a WAM, like a video world model~\citep{jang2025dreamgen,guo2026ctrlworld}, can also act as a data generator to synthesize additional training data, and the consistency between generated observations and actions has been used to filter such data with a learned classifier~\citep{kim2026robocurate}, to select rollouts at test time~\citep{ruan2026compatible}, and to terminate drifting rollouts during policy evaluation~\citep{tseng2026sc3eval}.
We extend this line to continual learning, regenerating previously learned tasks for data-free replay, and rectify the rollouts without training any additional model.

\section{Method}
\label{sec:method}
\begin{figure}[t]
 \centering
 \includegraphics[width=\linewidth]{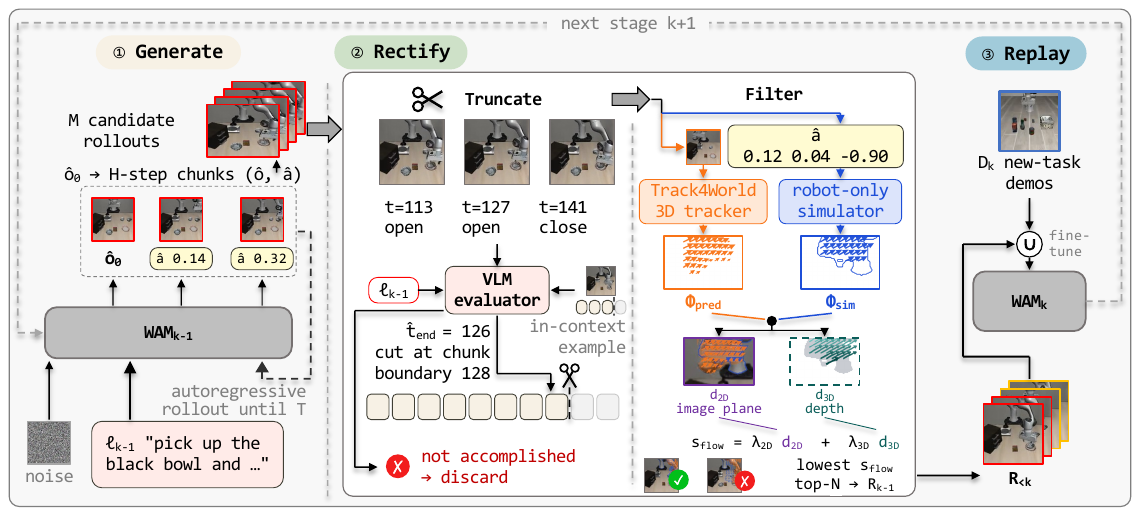}
 \caption{\textbf{Overview of \method.} After learning task $k-1$, $\mathrm{WAM}_{k-1}$ \ding{172} \textbf{generates} $M$ candidate rollouts of that task with the instruction $\ell_{k-1}$, which are then \ding{173} \textbf{rectified} by truncating each rollout at task completion and filtering out physically inconsistent ones. The top-$N$ retained rollouts $\mathcal{R}_{k-1}$ join those of earlier tasks to form the buffer $\mathcal{R}_{<k}$, which is \ding{174} \textbf{replayed} with the new-task demonstrations $\mathcal{D}_k$ when learning task $k$. The same procedure then repeats for the next stage $k+1$.}
 \label{fig:method}
\end{figure}

We present \method, which regenerates replay data for previously learned
tasks from their instructions alone and rectifies it before replay, as
illustrated in Fig.~\ref{fig:method}: the WAM generates candidate rollouts of
each task in Sec.~\ref{sec:generating-replay}, which are rectified by
completion-aware adaptive truncation and flow-based physical-consistency
filtering in Sec.~\ref{sec:rectified-replay} and replayed with new-task
demonstrations.

\subsection{Preliminaries}
\label{sec:continual-imitation-learning}

\paragraph{Continual learning in robotics.}
We model a robotic manipulation task \(\mathcal{T}_k\) by a task instruction
\(\ell_k\) and an expert demonstration dataset
\(\mathcal{D}_k=\{\tau_i^k\}_{i=1}^{|\mathcal{D}_k|}\). Each trajectory
\(\tau_i^k=\{(o_t,a_t)\}_{t=0}^{L_i^k-1}\) contains visual observations \(o_t\) and robot
actions \(a_t\), with \(L_i^k\) denoting its length. Continual learning
considers a robot that learns a sequence of tasks \(\{\mathcal{T}_k\}_{k=1}^{K}\) with a
single policy \(\pi_\theta(a_t\mid o_{\leq t},\ell_k)\) shared across all tasks and
conditioned on the instruction \(\ell_k\). At each stage, the policy is optimized by behavior
cloning on the demonstrations available for the current task:
\begin{equation}
 \mathcal{L}_{\mathrm{BC}}(\theta;\mathcal{D}_k)
 = \mathbb{E}_{\tau\sim\mathcal{D}_k}
 \left[\sum_{t=0}^{|\tau|-1}
 -\log \pi_\theta(a_t\mid o_{\leq t},\ell_k)\right].
 \label{eq:continual-bc}
\end{equation}
Sequential adaptation can degrade performance on previously learned tasks, resulting
in catastrophic forgetting. Experience replay (ER) mitigates this by maintaining a replay buffer
\(\mathcal{B}_{<k}\) that stores a subset of demonstrations from earlier
tasks, and optimizing the policy on \(\mathcal{D}_k\cup\mathcal{B}_{<k}\) at stage \(k\).
In contrast, we study the \emph{data-free} setting in which no data from previous
tasks is retained. After adapting to a task, only its instruction and the updated model parameters
carry over to later stages, while its demonstrations and task environment are no
longer accessible.

\label{sec:world-action-models}
\paragraph{World action models.}
Unlike a conventional policy that maps observations to actions, a WAM also
predicts the future observations that its actions would
produce~\citep{li2025uva,lingbotva2026,cosmospolicy2026,dreamzero2026}. Given a
task \(\mathcal{T}_k\) with instruction \(\ell_k\) and a trajectory prefix
\(\tau_{\leq t}\), it models the joint conditional distribution:
\begin{equation}
 p_\theta\bigl(
 o_{t+1:t+H},a_{t:t+H-1}
 \mid \tau_{\leq t},\ell_k
 \bigr),
 \label{eq:wam-joint-distribution}
\end{equation}
where \(H\) is the chunk size, i.e., the number of observations and actions predicted in one step.
In this paper, we adopt LingBot-VA~\citep{lingbotva2026} as a representative
WAM, which implements this formulation with a causal autoregressive
architecture that interleaves observation and action tokens.

\subsection{Generating Rollouts with WAM}
\label{sec:generating-replay}

We leverage the autoregressive rollout capability of WAMs to synthesize replay
data, generating a rollout for each learned task \(\mathcal{T}_j\) with its
instruction \(\ell_j\) as the only input. Built on a video generation model,
the WAM can generate observations from text alone, an ability we retain during
fine-tuning by keeping the initial observation as a prediction target
conditioned on the instruction. A rollout thus starts by sampling the initial
observation \(\hat{o}_0\sim p_\theta(o_0\mid\ell_j)\) and proceeds chunk by
chunk: at step \(t\), the WAM first predicts the next \(H\) observations from
the generated history \(\hat{\tau}_{\leq t}\), appends them to the model input,
and then predicts the corresponding actions conditioned on them:
\begin{equation}
 \hat{o}_{t+1:t+H}
 \sim p_\theta\bigl(\cdot\mid\hat{\tau}_{\leq t},\ell_j\bigr)
 \;\longrightarrow\;
 \hat{a}_{t:t+H-1}
 \sim p_\theta\bigl(\cdot\mid\hat{\tau}_{\leq t},\hat{o}_{t+1:t+H},\ell_j\bigr),
 \label{eq:rollout-step}
\end{equation}
which are appended to the model input as well. This repeats until a predefined
horizon \(T\), yielding the rollout
\(\hat{\tau}_j=\{(\hat{o}_t,\hat{a}_t)\}_{t=0}^{T-1}\).

\subsection{Rectified Generative Replay}
\label{sec:rectified-replay}

\paragraph{Completion-aware adaptive truncation.}
As discussed in Sec.~\ref{sec:generating-replay}, we generate rollouts with a
fixed horizon \(T\) for each task. Since the WAM is unable to recognize task
completion, it keeps generating after the task is done, producing purposeless
post-completion motion such as re-grasping the placed object or wandering
around. Such motion enters training as if it were part of the task, which
corrupts the learned behavior of the task. To remove it, we use a VLM as an agentic
evaluator to identify the completion point of each of the \(M\) candidate
rollouts generated for a task. Rather than feeding the whole video, we subsample candidate frames from the later part of the rollout and annotate each frame with its step index, gripper command, and end-effector speed from the generated actions, so that the evaluator can localize the completion precisely.
We further provide one in-context example from the same task, a generated rollout whose truncation range is annotated by a human (one per task), to calibrate its judgment. Given
the task instruction and these inputs, the evaluator determines whether the task
goal has been achieved and returns the end step \(\hat{t}_{\mathrm{end}}\),
defined as the point where the completing motion has fully finished, e.g., the
object has been released and the gripper has retreated. For tasks with multiple
sub-goals, the end step corresponds to the completion of the final sub-goal. We
then truncate the rollout at the chunk boundary following
\(\hat{t}_{\mathrm{end}}\), so that the completed interaction is retained while the motion after it is excluded. Rollouts judged as not accomplishing the task are discarded. The evaluator prompt and a truncation example are given in Fig.~\ref{fig:prompt} and Fig.~\ref{fig:truncation-example} in the appendix.

\paragraph{Flow-based physical-consistency filtering.}
The remaining issue is the observation--action drift caused by open-loop
generation, which we address by scoring each rollout's drift and filtering out
the worst. Measuring this drift requires knowing what actually happens when the generated actions are executed, which is unavailable since the environments of previous
tasks are no longer accessible. Fortunately, the robot itself is fully specified
by its URDF and camera parameters, so the motion of the robot under the
generated actions can be simulated without any environment and serves as a
reference. Concretely, we execute the actions of each truncated rollout
\(\tilde{\tau}\) in a robot-only simulator and record the resulting optical flow \(\Phi_{\mathrm{sim}}\) of the robot. In
parallel, we estimate the optical flow \(\Phi_{\mathrm{pred}}\) from the
generated observations with Track4World~\citep{lu2026track4world}, a dense 3D
tracker. Comparing the two flows on the robot pixels, we define the
physical-consistency score as:
\begin{equation}
 s_{\mathrm{flow}}(\tilde{\tau})
 =
 \lambda_{\mathrm{2D}}\,d_{\mathrm{2D}}
 \left(
 \Phi_{\mathrm{pred}},\Phi_{\mathrm{sim}}
 \right)
 +
 \lambda_{\mathrm{3D}}\,d_{\mathrm{3D}}
 \left(
 \Phi_{\mathrm{pred}},\Phi_{\mathrm{sim}}
 \right),
 \label{eq:flow-consistency}
\end{equation}
where \(d_{\mathrm{2D}}\) and \(d_{\mathrm{3D}}\) measure the discrepancy of the
two flows in the image plane and in 3D camera space, respectively, and lower scores indicate better agreement. Appendix~\ref{app:implementation} describes both terms and the weights \(\lambda_{\mathrm{2D}}\) and \(\lambda_{\mathrm{3D}}\). We rank the candidates by
\(s_{\mathrm{flow}}\) and retain the top-\(N\) as the replay set \(\mathcal{R}_j\)
for task \(\mathcal{T}_j\).

\paragraph{Replay with rectified rollouts.}
Finally, \method replays the rectified rollouts exactly as ER replays real
demonstrations in Sec.~\ref{sec:continual-imitation-learning}, replacing the
replay buffer \(\mathcal{B}_{<k}\) with the rectified rollouts
\(\mathcal{R}_{<k}=\bigcup_{j<k}\mathcal{R}_j\) and fine-tuning the WAM on
\(\mathcal{D}_k\cup\mathcal{R}_{<k}\) at stage \(k\) by maximizing the
likelihood of Eq.~(\ref{eq:wam-joint-distribution}).
Each \(\mathcal{R}_j\) is generated once and cached for all later stages, so \(\mathcal{R}_{<k}\) matches ER's \(\mathcal{B}_{<k}\) in size without any real data.

\section{Experiments}
\label{sec:exp}
\newcommand{\est}[1]{#1}

\subsection{Experiment Setup}
\label{sec:setup}

\paragraph{Benchmarks.}
\begin{wrapfigure}[15]{r}{0.3\linewidth}
\vspace{-\baselineskip}
\centering
\includegraphics[width=\linewidth]{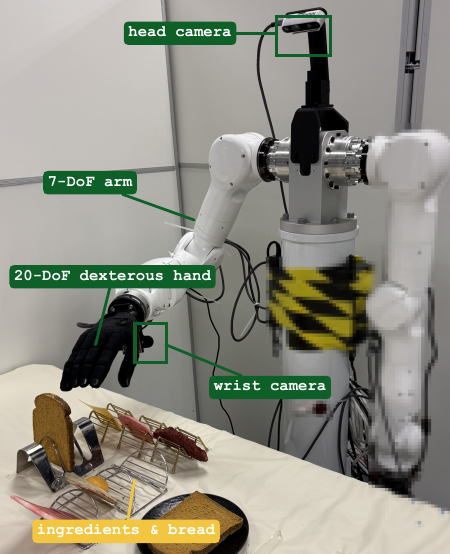}
\caption{\textbf{Real-robot setup.}}
\label{fig:real-setup}
\vspace{-\baselineskip}
\end{wrapfigure}
We evaluate continual learning in simulation on LIBERO and in the real world on five sequentially learned manipulation tasks. \textbf{For simulation}, we use LIBERO~\citep{liu2023libero}, a benchmark designed for continual learning in robotic manipulation. Its four suites, LIBERO-Spatial, LIBERO-Object, LIBERO-Goal, and LIBERO-10, each define a sequence of 10 tasks with 50 demonstrations per task. The four suites differ in what changes across tasks: the spatial layout, the manipulated object, the task goal, and long-horizon compositions of the three. \textbf{For the real world}, we design a sequence of five sandwich-making tasks on a robot with a 7-DoF arm and a 20-DoF dexterous hand, which observes the scene through a head camera and a wrist camera mounted on the hand (Fig.~\ref{fig:real-setup}). Each task is instructed as ``make a sandwich with $\langle$ingredient$\rangle$'', with the ingredient being steak, luncheon meat, cheese, a fried egg, or a meat slice in learning order, and requires the robot to grasp the ingredient, place it on the bread, grasp the top bread, and cover the sandwich (Fig.~\ref{fig:real-tasks}). We collect 50 teleoperated demonstrations per task; hardware and protocol details are given in Appendix~\ref{app:real-robot}.

\paragraph{Protocol and metrics.}
Following LIBERO, tasks are learned one after another, and after each task the model is evaluated on every task learned so far, which yields a lower-triangular task matrix. Tasks in every LIBERO suite are ordered lexicographically by their instructions, and the five real-robot tasks follow a fixed order shared by all methods; the full sequences are listed in Appendix~\ref{app:task-order}. In simulation, each task is evaluated for 50 trials and we record its success rate as defined by LIBERO. On the real robot, a binary success hides partial progress on these long-horizon tasks, so we report the progress rate instead: each trial is scored by how many of the four steps it completes, divided by four (Appendix~\ref{app:real-robot}). Let $c_{i,k}$ denote the performance of task $i$ after learning task $k$, averaged over its evaluation trials: the success rate over 50 trials in simulation, or the progress rate over 20 trials on the real robot. Let $K$ denote the number of tasks. We compute three continual learning metrics from the task matrix:
\begin{equation}\small
\mathrm{SR}=\tfrac{1}{K}\sum\nolimits_{i}c_{i,K},\quad
\mathrm{NBT}=\tfrac{1}{K}\sum\nolimits_{k<K}\tfrac{1}{K-k}\sum\nolimits_{j>k}(c_{k,k}-c_{k,j}),\quad
\mathrm{Plasticity}=\tfrac{1}{K}\sum\nolimits_{k}c_{k,k}.
\label{eq:metrics}
\end{equation}
SR is the average performance over all tasks evaluated with the final checkpoint, i.e., after learning the last task; NBT (negative backward transfer)~\citep{liu2023libero} measures forgetting as the average drop of each task from the checkpoint that has just learned it to all later checkpoints; Plasticity is the average performance of each task at the checkpoint that has just learned it. Higher SR and Plasticity and lower NBT are better.

\paragraph{Baselines.}
All methods share the same LingBot-VA backbone~\citep{lingbotva2026}, training schedule, and evaluation protocol. We consider two groups of baselines. The first group does not replay: sequential fine-tuning (Seq-FT), sequential fine-tuning of LoRA adapters~\citep{hu2022lora} (Seq-LoRA), EWC~\citep{kirkpatrick2017ewc}, and LwF~\citep{li2017lwf}. EWC constrains parameter updates with a quadratic penalty weighted by the Fisher importance of each parameter to previous tasks, and LwF regularizes the model by matching its predictions on new-task inputs to those of the previous model. The second group replays 20 episodes per previous task: experience replay (ER)~\citep{rolnick2019er} replays real demonstrations and serves as an upper bound, while R-WAM replays the generated rollouts without rectification. R-WAM keeps all $N=20$ rollouts it generates, whereas \method selects the same number from $M=40$ candidates.

\paragraph{Implementation.}
After learning each task, \method generates $M=40$ candidate rollouts from the instruction alone, truncates them with a VLM evaluator, and retains the $N=20$ rollouts with the best physical-consistency scores (all experiments use $M=2N$), computed with Track4World~\citep{lu2026track4world} flow against a robot-only MuJoCo simulation. The retained rollouts are generated once and reused in all later stages. Replay data and new-task demonstrations are mixed with equal sampling weight. Training hyperparameters are given in Appendix~\ref{app:setup}; generation horizons and the evaluator prompt are given in Appendix~\ref{app:implementation}.

\subsection{Effectiveness of Rectified Generative Replay}
\label{sec:main-results}

\begin{table}[t]
\centering
\caption{\textbf{Continual learning results on the four LIBERO suites.} We compare methods without replay (top) and with replay (bottom); all replay methods use 20 episodes per previous task. ER (gray, italic) replays real demonstrations and serves as an upper bound. R-WAM$^\dagger$ replays generated rollouts without rectification. $\pm$ denotes the standard deviation over three training seeds. Plast.\ denotes Plasticity; all metrics follow Section~\ref{sec:setup}.}
\label{tab:main}
\vspace{2pt}
\scriptsize
\setlength{\tabcolsep}{2.2pt}
\resizebox{\linewidth}{!}{
\begin{tabular}{@{}l*{15}{c}@{}}
\toprule
\textbf{Method}
& \multicolumn{3}{c}{LIBERO-Spatial}
& \multicolumn{3}{c}{LIBERO-Object}
& \multicolumn{3}{c}{LIBERO-Goal}
& \multicolumn{3}{c}{LIBERO-10}
& \multicolumn{3}{c}{Average} \\
\cmidrule(lr){2-4}\cmidrule(lr){5-7}\cmidrule(lr){8-10}\cmidrule(lr){11-13}\cmidrule(lr){14-16}
& SR $\uparrow$ & NBT $\downarrow$ & Plast. $\uparrow$
& SR $\uparrow$ & NBT $\downarrow$ & Plast. $\uparrow$
& SR $\uparrow$ & NBT $\downarrow$ & Plast. $\uparrow$
& SR $\uparrow$ & NBT $\downarrow$ & Plast. $\uparrow$
& SR $\uparrow$ & NBT $\downarrow$ & Plast. $\uparrow$ \\
\midrule
Seq-FT & 11.2 & 86.07 & 95.8 & 21.2 & 67.84 & 97.4 & 9.8 & 84.87 & 94.8 & 7.6 & 82.00 & 89.6 & 12.5 & 80.19 & 94.4 \\
Seq-LoRA & 9.6 & 75.42 & 86.2 & 17.6 & 52.47 & 87.0 & 9.8 & 68.72 & 84.2 & 7.4 & 71.24 & 80.4 & 11.1 & 66.96 & 84.4 \\
EWC & 12.0 & 85.59 & 96.2 & 23.6 & 62.19 & 98.0 & 10.0 & 84.55 & 95.2 & 7.8 & 82.24 & 90.6 & 13.3 & 78.64 & 95.0 \\
LwF & 13.0 & 85.14 & 96.0 & 27.6 & 61.76 & 96.8 & 10.4 & 83.70 & 95.0 & 8.6 & 78.96 & 88.8 & 14.9 & 77.39 & 94.2 \\
\midrule
\rowcolor{gray!12}
\multicolumn{1}{@{}>{\columncolor{gray!12}[0pt][\tabcolsep]}l}{\textit{ER}} & \textit{94.0} & \textit{0.54} & \textit{94.3} & \textit{97.3} & \textit{$-$1.83} & \textit{95.5} & \textit{92.8} & \textit{2.11} & \textit{95.7} & \textit{94.1} & \textit{$-$0.46} & \textit{92.7} & \textit{94.6} & \textit{0.09} & \multicolumn{1}{>{\columncolor{gray!12}[\tabcolsep][0pt]}c@{}}{\textit{94.6}} \\[-1.5pt]
\rowcolor{gray!12}
\multicolumn{1}{@{}>{\columncolor{gray!12}[0pt][\tabcolsep]}l}{} & {\tiny\color{gray}$\pm$0.3} & {\tiny\color{gray}$\pm$0.60} & {\tiny\color{gray}$\pm$1.1} & {\tiny\color{gray}$\pm$0.8} & {\tiny\color{gray}$\pm$1.16} & {\tiny\color{gray}$\pm$1.7} & {\tiny\color{gray}$\pm$2.8} & {\tiny\color{gray}$\pm$1.45} & {\tiny\color{gray}$\pm$0.9} & {\tiny\color{gray}$\pm$0.3} & {\tiny\color{gray}$\pm$0.43} & {\tiny\color{gray}$\pm$0.3} & {\tiny\color{gray}$\pm$0.5} & {\tiny\color{gray}$\pm$0.79} & \multicolumn{1}{>{\columncolor{gray!12}[\tabcolsep][0pt]}c@{}}{{\tiny\color{gray}$\pm$0.7}} \\
R-WAM$^\dagger$ & 76.1 & 11.97 & 93.6 & 83.1 & 18.93 & 94.2 & 51.2 & 27.31 & 83.3 & 71.3 & 7.63 & 85.1 & 70.4 & 16.46 & 89.1 \\[-1.5pt]
& {\tiny\color{gray}$\pm$8.7} & {\tiny\color{gray}$\pm$2.58} & {\tiny\color{gray}$\pm$4.0} & {\tiny\color{gray}$\pm$6.8} & {\tiny\color{gray}$\pm$5.93} & {\tiny\color{gray}$\pm$1.4} & {\tiny\color{gray}$\pm$4.9} & {\tiny\color{gray}$\pm$1.75} & {\tiny\color{gray}$\pm$3.6} & {\tiny\color{gray}$\pm$2.6} & {\tiny\color{gray}$\pm$2.69} & {\tiny\color{gray}$\pm$4.4} & {\tiny\color{gray}$\pm$0.4} & {\tiny\color{gray}$\pm$1.13} & {\tiny\color{gray}$\pm$1.1} \\
\method (ours) & 94.0 & 2.90 & 95.7 & 95.5 & $-$0.31 & 95.3 & 92.2 & 2.00 & 95.5 & 88.9 & 4.73 & 92.0 & 92.7 & 2.33 & 94.7 \\[-1.5pt]
& {\tiny\color{gray}$\pm$1.1} & {\tiny\color{gray}$\pm$0.32} & {\tiny\color{gray}$\pm$0.8} & {\tiny\color{gray}$\pm$0.4} & {\tiny\color{gray}$\pm$1.05} & {\tiny\color{gray}$\pm$0.8} & {\tiny\color{gray}$\pm$2.3} & {\tiny\color{gray}$\pm$3.17} & {\tiny\color{gray}$\pm$3.1} & {\tiny\color{gray}$\pm$4.2} & {\tiny\color{gray}$\pm$2.28} & {\tiny\color{gray}$\pm$1.8} & {\tiny\color{gray}$\pm$1.2} & {\tiny\color{gray}$\pm$1.08} & {\tiny\color{gray}$\pm$0.4} \\
\bottomrule
\end{tabular}
}
\end{table}

\paragraph{\method nearly matches ER without any data from previous tasks.}
Table~\ref{tab:main} compares \method with the no-replay and replay baselines on the four LIBERO suites; all replay methods use the same buffer size of 20 episodes per previous task. We draw three observations. \textbf{First}, without replay, the WAM forgets almost everything it has learned: Seq-FT, Seq-LoRA, EWC, and LwF reach an average SR of only 11--15\% with an NBT of 67--80, although their Plasticity stays above 84\%. \textbf{Second}, without any real data from previous tasks, \method reaches an average SR of 92.7\% and an NBT of 2.33, only 1.9~pp below ER on average, which serves as an upper bound since it replays real demonstrations of previous tasks. The gap is largest on LIBERO-10 ( 5.2~pp), consistent with its long-horizon tasks, where observation--action drift may accumulate over longer rollouts. \textbf{Finally}, replaying raw rollouts substantially reduces forgetting relative to Seq-FT, raising the average SR from 12.5\% to 70.4\%, but still lags ER by a wide margin, which shows the necessity of rectification.

\begin{table}[t]
\begin{minipage}[t]{0.47\linewidth}
\centering
\caption{\textbf{Real-world results on five sequential tasks.} Metrics are computed on the progress rate (Appendix~\ref{app:real-robot}). ER (gray, italic) serves as an upper bound.}
\label{tab:real_robot}
\small
\setlength{\tabcolsep}{9.5pt}
\renewcommand{\arraystretch}{1.072}
\begin{tabular}{@{}lccc@{}}
\toprule
\textbf{Method} & SR $\uparrow$ & NBT $\downarrow$ & Plast.\ $\uparrow$ \\
\midrule
\rowcolor{gray!12}
\multicolumn{1}{@{}>{\columncolor{gray!12}[0pt][\tabcolsep]}l}{\textit{ER}} & \textit{91.8} & \textit{2.27} & \multicolumn{1}{>{\columncolor{gray!12}[\tabcolsep][0pt]}c@{}}{\textit{94.8}} \\
Seq-FT & 28.5 & 65.62 & \textbf{98.8} \\
\method (ours) & \textbf{89.5} & \textbf{3.45} & 93.8 \\
\bottomrule
\end{tabular}
\end{minipage}\hfill
\begin{minipage}[t]{0.5\linewidth}
\centering
\caption{\textbf{Ablation of \method components} on LIBERO-Spatial. Each row removes one rectification component, and R-WAM applies none.}
\label{tab:ablation}
\small
\begin{tabular*}{\linewidth}{@{\extracolsep{\fill}}lccc@{}}
\toprule
\textbf{Variant} & SR $\uparrow$ & NBT $\downarrow$ & Plast.\ $\uparrow$ \\
\midrule
R-WAM (no rectification) & 76.1 & 11.97 & 93.6 \\
\method w/o truncation & 70.8 & 13.73 & 89.2 \\
\method w/o filtering & 80.8 & 7.02 & 91.4 \\
\method (ours) & \textbf{94.0} & \textbf{2.90} & \textbf{95.7} \\
\bottomrule
\end{tabular*}
\end{minipage}
\end{table}

\paragraph{Real-world results.}
\looseness=-1
Table~\ref{tab:real_robot} compares \method with Seq-FT and ER on the five real-world tasks, where the trend matches simulation: Seq-FT suffers catastrophic forgetting (28.5\% SR, 65.62 NBT), whereas \method retains previous tasks nearly as well as ER (89.5\% versus 91.8\% SR, 3.45 versus 2.27 NBT) without any real data from previous tasks. Both replay methods also give up 4--5~pp of Plasticity relative to Seq-FT, which we attribute to interference among co-trained tasks~\citep{yu2020gradient,wu2025discrete}, here between tasks that differ only in a visually similar target. For more case studies and qualitative results, see Appendix~\ref{app:real-qual} and the supplementary video.

\subsection{Ablation Study on \method}
\label{sec:ablation}

\paragraph{Both components are necessary.}
Table~\ref{tab:ablation} removes each rectification component on LIBERO-Spatial. \textbf{First}, removing either component degrades performance, from 94.0\% SR to 80.8\% without filtering and to 70.8\% without truncation, so both are necessary. \textbf{Second}, removing truncation performs no better than R-WAM (70.8\% vs.\ 76.1\%), which replays the raw rollouts, suggesting that the purposeless motion at the end of a rollout may interfere with the consistency score and thus make~filtering unreliable.

\paragraph{Ablation of completion-aware adaptive truncation.}
\begin{wraptable}{r}{0.46\linewidth}
\vspace{-\dimexpr\intextsep+0.4\baselineskip\relax}
\caption{\textbf{Ablation of completion-aware adaptive truncation.} Accuracy is the fraction of predicted truncations that fall inside the human-annotated range. We ablate the VLM (top) and the prompt design (bottom); Terra with the full prompt is our default.}
\label{tab:evaluator}
\centering
\scriptsize
\begin{tabular*}{\linewidth}{@{\extracolsep{\fill}}llc@{}}
\toprule
& \textbf{Setting} & \textbf{Acc.\ (\%) $\uparrow$} \\
\midrule
\multirow{4}{*}{\rotatebox{90}{\scriptsize VLM}}
& GPT-5.6 Sol~\citep{openai2026gpt56} & \textbf{93.5} \\
& GPT-5.6 Terra & 91.0 \\
& GPT-5.6 Luna & 88.0 \\
& Qwen3-VL~\citep{bai2025qwen3vl} & 73.5 \\
\midrule
\multirow{3}{*}{\rotatebox{90}{\scriptsize Prompt}}
& Full prompt & \textbf{91.0} \\
& w/o in-context example & 78.0 \\
& w/o action annotations & 76.0 \\
\bottomrule
\end{tabular*}
\vspace{-\intextsep}
\end{wraptable}
To examine adaptive truncation itself, we sample five generated rollouts from each of the 40 LIBERO tasks, annotate an acceptable range of truncation steps for each of the 200 samples, and count a predicted truncation as correct if it falls inside the range. Table~\ref{tab:evaluator} ablates the VLM evaluator and its prompt on this set. \textbf{First}, stronger VLMs truncate more accurately, from 73.5\% for Qwen3-VL to 93.5\% for GPT-5.6 Sol; we use Terra (91.0\%) in all other experiments to balance accuracy against cost. \textbf{Second}, removing the in-context example lowers accuracy to 78.0\%, likely because where a task ends is inherently ambiguous and the example anchors the VLM's notion of completion. \textbf{Finally}, removing the action annotations lowers accuracy to 76.0\%, suggesting that the moment of completion is often hard to tell from observations alone and the motion cues of the actions help.

\begin{figure}[t]
\centering
\includegraphics[width=0.96\linewidth]{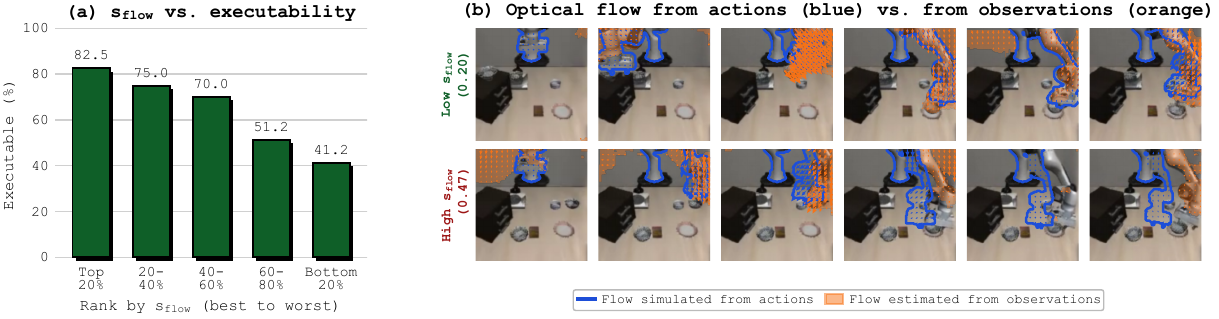}
\caption{\textbf{What the physical-consistency score measures.} \textbf{(a)} We use the WAM to generate 400 rollouts on LIBERO-Spatial, group them by $s_{\mathrm{flow}}$ from most to least consistent, and execute their actions in simulation. Each bar is the fraction of rollouts in that group whose actions complete the task. \textbf{(b)} Optical flows behind $s_{\mathrm{flow}}$ for a low-score and a high-score rollout: blue is simulated from the generated actions, orange is estimated from the generated observations.}
\label{fig:consistency}
\end{figure}

\paragraph{The physical-consistency score $s_{\mathrm{flow}}$ identifies executable rollouts.}
To verify this, we use the WAM to generate 400 rollouts on LIBERO-Spatial, starting from real initial observations so that their actions can be executed open-loop in simulation. We regard a rollout as executable if its execution accomplishes the task, and report the executable fraction of rollouts at different levels of $s_{\mathrm{flow}}$ in Fig.~\ref{fig:consistency}(a). The fraction decreases steadily from 82.5\% for the most consistent 20\% to 41.2\% for the least consistent 20\%, showing that $s_{\mathrm{flow}}$ indeed reflects whether the generated actions agree with the generated observations, and filtering by it keeps the executable rollouts. In Fig.~\ref{fig:consistency}(b), we further visualize the optical flows behind $s_{\mathrm{flow}}$, where blue is simulated from the generated actions and orange is estimated from the generated observations. In a low-$s_{\mathrm{flow}}$ rollout the two flows point in the same direction throughout, whereas in a high-$s_{\mathrm{flow}}$ rollout they diverge over time and gradually separate, which eventually leads to observation--action drift.

\begin{figure}[t]
\begin{minipage}[t]{0.46\linewidth}
\centering
\includegraphics[width=\linewidth]{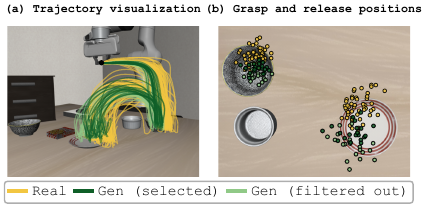}
\captionof{figure}{\textbf{Visualization of real and generated trajectories on LIBERO.} \textbf{(a)} Trajectories of 50 real demonstrations and 40 generated rollouts on a LIBERO-Spatial task, 20 selected by \method and 20 filtered out. \textbf{(b)} Top view of where each trajectory grasps the bowl and releases it on the plate.}
\label{fig:traj-overlay}
\end{minipage}\hfill
\begin{minipage}[t]{0.52\linewidth}
\centering
\includegraphics[width=\linewidth]{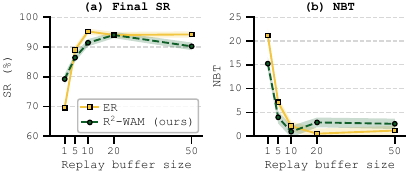}
\captionof{figure}{\textbf{Scaling the replay buffer on LIBERO-Spatial.} Final SR and NBT as the replay buffer grows from 1 to 50 episodes per previous task. ER replays real demonstrations of previous tasks, whereas \method replays generated rollouts and uses no real data at all. Shading denotes the standard error over 50 trials per task.}
\label{fig:budget-curves}
\end{minipage}
\end{figure}

\paragraph{Rectified rollouts overlap with real demonstrations but do not fully match them.}
To compare the replay data of ER and \method more directly, we visualize the trajectories of the real demonstrations and the generated rollouts of a LIBERO-Spatial task in Fig.~\ref{fig:traj-overlay}, where the 40 generated rollouts are all adaptively truncated and then split by the physical-consistency filtering into 20 selected and 20 filtered out. We draw three observations from the comparison. \textbf{First}, the generated trajectories follow the same pattern as the real ones in Fig.~\ref{fig:traj-overlay}(a), but they gradually deviate from the real ones as the rollout proceeds, which is evident in Fig.~\ref{fig:traj-overlay}(b), where the release positions deviate much more than the grasp positions. Since the generated videos are visually indistinguishable from real ones, this deviation reflects the observation--action drift accumulating over open-loop generation. \textbf{Second}, the physical-consistency score identifies this drift well, as the selected rollouts stay much closer to the real trajectories than the filtered-out ones. This also explains why the higher-ranked rollouts in Fig.~\ref{fig:consistency}(a) are more often executable. \textbf{Finally}, filtering only mitigates the drift rather than removing it, as some deviation remains in the selected rollouts, which is consistent with the residual gap to ER that remains on the longer-horizon suite.

\paragraph{Generated replay tracks real replay across buffer sizes.}
To examine the effect of the replay buffer size, we scale it from 1 to 50 episodes per task on LIBERO-Spatial and compare \method with ER at every size, as shown in Fig.~\ref{fig:budget-curves}. \textbf{First}, both methods improve rapidly with the buffer size and no longer improve beyond 20 episodes, so we use a buffer of 20 in the main experiments to balance performance against generation cost. \textbf{Second}, \method even outperforms ER at a buffer size of 1 (79.2\% versus 69.6\% SR), though ER overtakes it quickly and stays slightly ahead. We conjecture that a generated rollout reflects the typical behavior the model has learned for a task and thus anchors it better than a single real demonstration, which may be atypical or noisy. With more real demonstrations, however, ER covers the true distribution, while generated rollouts add little new coverage and still carry some residual drift even after rectification. The task matrices behind Fig.~\ref{fig:budget-curves} are given in Fig.~\ref{fig:budget} in the appendix.

\subsection{Diagnostic Study: Forgetting Concentrates in Inverse Dynamics}
\label{sec:analysis}
\begin{figure}[t]
\centering
\includegraphics[width=0.96\linewidth]{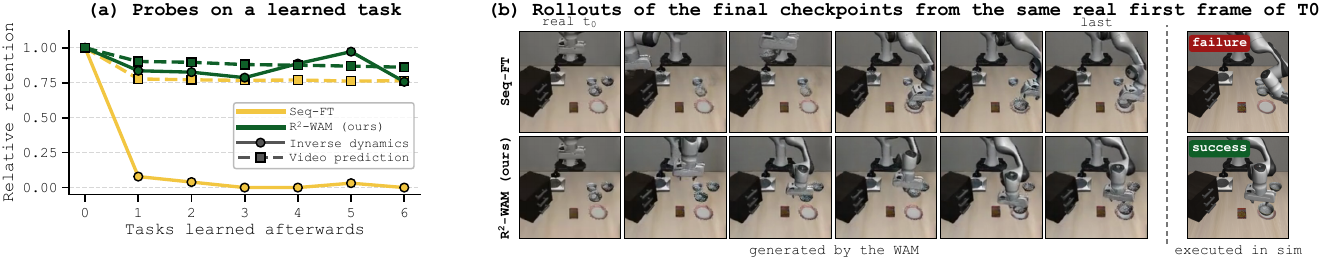}
\caption{\textbf{Which capabilities of a WAM are forgotten.} \textbf{(a)} Relative retention $r(\Delta)$ (Eq.~\ref{eq:retention}) of two probed capabilities, $\Delta$ tasks after a task was learned. Circles with solid lines: inverse dynamics modeling (actions from real observations); squares with dashed lines: video prediction (observations from real actions). Yellow: Seq-FT; green: \method. \textbf{(b)} Rollouts generated by the final checkpoints from the same real first frame of T0, with the \method rollout shown after truncation. The rightmost frame is rendered by executing the generated actions in simulation, which reveals whether they accomplish the task.}
\label{fig:probes}
\end{figure}

A WAM couples two capabilities, inverse dynamics modeling and video prediction, and we ask which of them is lost during continual learning. We probe each capability separately on LIBERO-Spatial with teacher forcing. For inverse dynamics modeling, the model receives real observations of a task and predicts the actions, which we execute in simulation and score by success rate. For video prediction, it receives the real actions and predicts the observations, scored by PSNR, which measures how closely the predicted frames match the real ones. Let $p_{i,k}$ be the score of a probe on task $i$ after learning task $k$ ($k \ge i$), over 20 samples. We measure forgetting by relative retention
\begin{equation}
\small
r(\Delta) = \frac{1}{K-\Delta}\sum\nolimits_{i=0}^{K-1-\Delta} \frac{p_{i,i+\Delta}}{p_{i,i}},
\label{eq:retention}
\end{equation}
i.e., the fraction of a task's initial score that remains $\Delta$ tasks later, averaged over tasks, where lower values mean more forgetting.

Fig.~\ref{fig:probes} reports the results, from which we draw three observations. \textbf{First}, Seq-FT forgets inverse dynamics modeling almost entirely: after learning a new task, the retention of this capability drops to 8\% and never recovers, whereas under \method it stays above 75\% across all subsequent tasks. \textbf{Second}, video prediction is largely preserved, retaining 78\% under Seq-FT and over 85\% under \method, so even a WAM whose actions have collapsed still imagines visually plausible observations. \textbf{Finally}, Fig.~\ref{fig:probes}(b) visualizes rollouts generated by the final checkpoints of Seq-FT and \method from the same real first frame of T0: both produce a fluent video of the task, but only the actions of \method accomplish it when executed in simulation.

\section{Conclusion}
\label{sec:conclusion}
We presented \method, a data-free approach to continual learning for world action models. A WAM can regenerate rollouts of previously learned tasks from their instructions alone, but such rollouts cannot be replayed directly for two reasons: the WAM does not stop generating at task completion, and its generated observations gradually drift from the generated actions. \method addresses both issues with rectified generative replay, which truncates each rollout at task completion and filters out physically inconsistent ones before replay. On the four LIBERO suites and five real-world tasks, \method reduces forgetting by over 90\% relative to sequential fine-tuning and nearly reaches experience replay without retaining any data from previous tasks. Ablations confirm that both truncation and filtering are necessary, and our analysis shows that forgetting in a WAM concentrates in inverse dynamics modeling while video prediction is largely preserved. This explains why a WAM can regenerate its past for replay and points to the action side as the part to verify.

\paragraph{Limitations.}
\looseness=-1
Two limitations point to future work. First, a small gap to ER remains and is largest on long-horizon tasks, suggesting that filtering mitigates but does not fully resolve observation--action drift, and aligning the two at training time or grounding generation with an external world model are directions worth exploring. Second, the reduced plasticity of replay-based methods in our real-world experiments suggests that interference between tasks also needs to be addressed.

\subsection*{AI use statement}
In this work, we used generative AI tools for two tasks that require disclosure: implementing methods (assistants wrote code for the proposed method and the experimental pipeline under the authors' specifications) and cleaning and reformatting data (converting training data into the format required by the training pipeline, and aggregating evaluation results into the task matrices and tables reported in the paper). We have not used generative AI tools for the other tasks with required disclosure, and the remaining ones are not applicable to this work. Additionally, we used generative AI tools for drafting and editing the paper, polishing grammar and wording, producing plots and illustrations, identifying relevant literature, and formatting references. We have reviewed all AI-assisted work: all code was verified and tested by the authors, and all reported numbers were traced back to the experimental logs. We take responsibility for the final content of this work, including text, claims, or artifacts produced with the aid of generative AI.

\subsection*{Reproducibility statement}
Section~\ref{sec:setup} and Appendix~\ref{sec:appendix-placeholder} give the full experimental protocol, including training details, the evaluator prompt, and task sequences. A supplementary video shows the real-robot experiments. Code will be released upon publication.

\bibliography{references}

\newpage
\appendix
\section{Appendix}
\label{sec:appendix-placeholder}
\raggedbottom

\paragraph{Supplementary video.}
To better understand our work, we supplement an intuitive introduction of \method and real-robot experimental results in \texttt{\textbf{supplementary\_video.mp4}}.

\subsection{Experimental Setup}
\label{app:setup}

\paragraph{Training and evaluation.}
Every task is trained for 1{,}000 steps with a global batch size of 64 and a learning rate of $2\times10^{-5}$, saving a checkpoint every 250 steps; the checkpoint with the best success on the current task over 16 validation trials is kept, using initial states disjoint from the 50 test trials. Replay data and new-task demonstrations are mixed with equal per-sample weight. All LIBERO methods use the same set of training seeds and the same evaluation scenes.

\paragraph{Backbone and generation.}
LingBot-VA couples a Wan-family video diffusion transformer with an action
branch: a 5.1B-parameter DiT (30 layers, hidden size 3072, FFN 14336) over
48-channel latents produced by a Wan VAE, conditioned on a frozen UMT5-XXL text
encoder. On LIBERO it observes two $128\times128$ views and predicts four latent
frames per chunk, i.e.\ $H=16$ actions, which is also the granularity at which
truncation cuts. Generation is autoregressive across chunks and diffusion-based
within a chunk, and we keep the released inference configuration unchanged: a
flow-matching sampler with 20 denoising steps for observations and 50 for
actions, and classifier-free guidance 5.0 on observations and 1.0 on actions.
The one component we change is action normalization: each task's actions are
normalized by their own 1st and 99th percentiles instead of a single global
range.

\begin{figure}[t]
\centering
\includegraphics[width=0.92\linewidth]{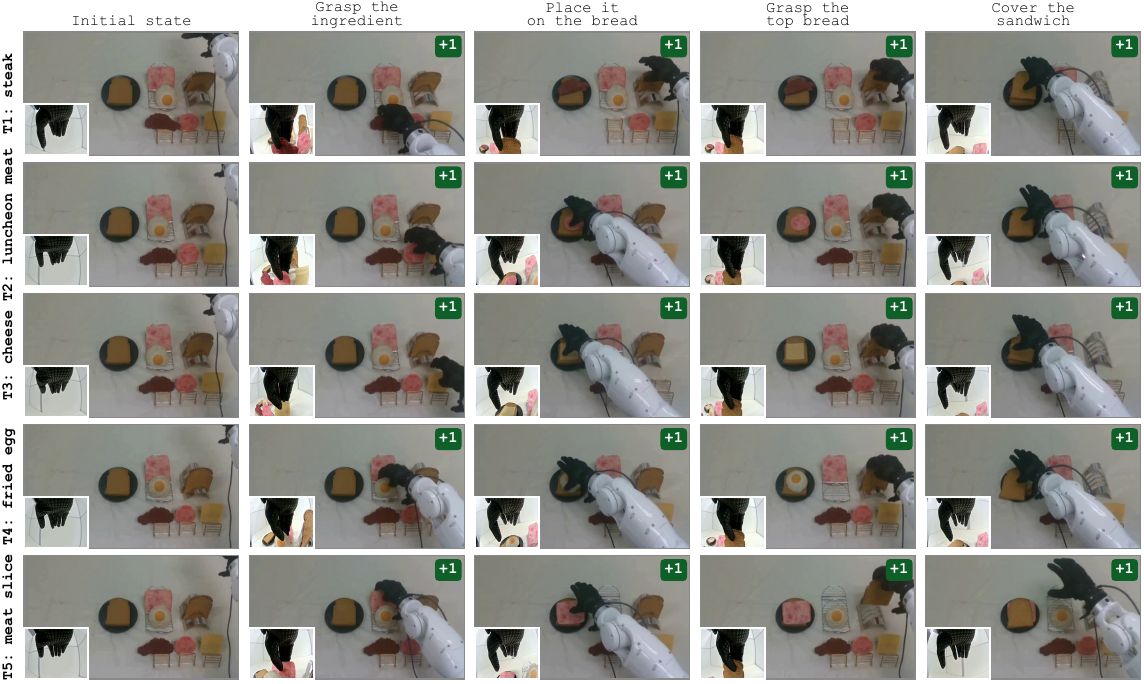}
\caption{\textbf{The five real-world sandwich tasks and their scored steps.} Each row shows one demonstration of a task from the head camera, with the wrist camera inset at the lower left; columns show the initial state and the four scored steps, each worth one point of the four-point progress score. Tasks are learned in the order T1 to T5 and share the same scene, so a policy must attend to the instruction to pick the right ingredient.}
\label{fig:real-tasks}
\end{figure}

\label{app:real-robot}
\paragraph{Real-robot setup.}
\textbf{\ding{172} Platform and data.} As shown in Fig.~\ref{fig:real-setup}, the robot is a 7-DoF arm with a 20-DoF dexterous hand, observed by a head camera and a wrist camera. Observations are recorded at 30\,Hz and resized to $256\times320$, and the action is the 27-dimensional joint command. We fine-tune the WAM on the 50 demonstrations of each task with the same schedule as in LIBERO. \textbf{\ding{173} Tasks.} The five tasks share a tabletop with a plate holding a slice of bread, a rack of five ingredients, and a stack of bread, as illustrated in Fig.~\ref{fig:real-tasks}. They differ only in the instructed ingredient and the grasp it demands, so forgetting manifests as picking the wrong ingredient or failing the grasp. \textbf{\ding{174} Scoring.} Each trial is scored on four steps independently, one point each: the instructed ingredient is lifted off the rack, it rests on the bread, the top bread is lifted, and the sandwich is covered. The first two steps earn no point if a wrong ingredient is used, whereas the last two are identical across tasks, so a policy that has forgotten the ingredient can still earn up to half the points. The progress rate is the score divided by four, averaged over 20 trials per task, and SR, NBT, and Plasticity are all computed on it in place of the binary success.

\subsection{Details of Rectified Generative Replay}
\label{app:implementation}

\paragraph{Generation.}
For each learned task, \method samples 40 candidate rollouts from the instruction alone. The generation horizon of a task is set to $\lfloor \bar{L}/16 \rfloor + 2$ action chunks of 16 steps, where $\bar{L}$ is the mean length of its demonstrations, i.e., about two chunks longer than a typical demonstration; the resulting ranges for the four suites are given in Table~\ref{tab:horizons}. We use the demonstration length only as a convenient way to set this horizon, and it carries no information about the task beyond a rough duration: in practice the horizon can equally be set from a human estimate of how long the task takes, or by probing the model with a few generated rollouts of the task before replay. Moreover, because completion-aware truncation cuts every rollout at task completion, the horizon only needs to be long enough, and its exact value has little effect on the replay data.

\begin{table}[h]
\centering
\caption{Demonstration lengths and generation horizons per suite (range over tasks, in steps).}
\label{tab:horizons}
\small
\begin{tabular}{@{}lcccc@{}}
\toprule
& Spatial & Object & Goal & LIBERO-10 \\
\midrule
Mean demonstration length & 101--150 & 133--161 & 89--204 & 189--416 \\
Generation horizon & 128--176 & 160--192 & 112--224 & 208--432 \\
\bottomrule
\end{tabular}
\end{table}

\begin{figure}[t]
\centering
\includegraphics[width=0.95\linewidth]{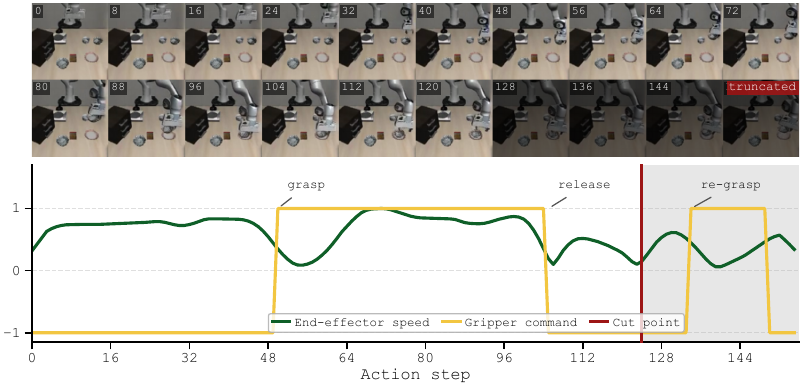}
\caption{\textbf{A truncated rollout for the task ``put the bowl on the plate''.} Top: frames every 8 steps; shaded frames are removed. Bottom: normalized end-effector speed and gripper command, with the cut point of the evaluator. The placement, the release, and the upward retreat are retained, while the subsequent descent and re-grasping loop are removed.}
\label{fig:truncation-example}
\end{figure}

\paragraph{Completion-aware adaptive truncation.}
The evaluator is a VLM queried with the task instruction and two parts of visual evidence, with the full prompt given in Fig.~\ref{fig:prompt} of Appendix~\ref{app:prompt}. Part 1 is an in-context example: a generated rollout of the same task in which a human has marked the range of acceptable truncation steps, so that the VLM learns what completion looks like for this task. Part 2 is the candidate rollout to be judged. Instead of the whole video, we uniformly sample frames from the last part of the rollout, every 2 steps over the final 20\% or at least 40 steps. Frames at which a motion ends or the gripper releases, as detected from the generated actions, are tagged \texttt{[BOUNDARY]} and the others \texttt{[SAMPLE]}, since the correct truncation usually falls at a boundary. Every frame pairs the external and the wrist view of one step, and we additionally attach its step index, gripper command, and end-effector velocity, along with a summary of the motion segments detected from the generated actions. Fig.~\ref{fig:truncation-example} shows an example, where the placement, the release, and the upward retreat are retained while the subsequent re-grasping loop is removed.
\label{app:truncation}

\paragraph{Physical-consistency score.}
Both flows are computed between frames of the truncated rollout: \(\Phi_{\mathrm{sim}}\) by rendering the robot in a robot-only MuJoCo scene driven by the generated actions, and \(\Phi_{\mathrm{pred}}\) by Track4World on the generated video. The comparison is restricted to robot pixels. A pixel is included only if the simulator marks it as part of the moving robot in both frames of the pair. The score therefore measures whether the generated video moves the robot the way the generated actions would, and task objects are not scored. \(d_{\mathrm{2D}}\) and \(d_{\mathrm{3D}}\) summarize the disagreement between the two flows in the image plane and in 3D camera space, respectively, as a normalized discrepancy in \([0,1]\) averaged over all robot pixels and frame pairs of the rollout. We set \(\lambda_{\mathrm{2D}}=\lambda_{\mathrm{3D}}=0.5\) and use them unchanged for all suites and the real robot.

\paragraph{Real-robot variant.}
For the dexterous hand, the gripper command shown to the evaluator is replaced by the mean distance between the thumb and the other fingertips, mapped to the same open/close scale, and the end-effector velocity is computed from the arm joint commands by forward kinematics; the flow filter uses the head camera only. The prompt is otherwise unchanged.

\paragraph{Cost of rectified generative replay.}
Table~\ref{tab:cost} lists the wall-clock time of each stage for one LIBERO-Spatial task on a single node with four H200 GPUs, averaged over the tasks of the chain behind Table~\ref{tab:main}. Generating and rectifying the replay set of a task takes about 18 minutes, roughly one sixth of fine-tuning on that task, and is incurred once per task since the rectified rollouts are reused in all later stages. Generation scales linearly with the number of candidates $M$ and with the horizon; truncation is bounded by the VLM API rather than by compute.

\begin{table}[t]
\centering
\caption{Per-task cost of \method on LIBERO-Spatial (4$\times$H200). Rectification stages are measured on $M=40$ candidate rollouts of 128--176 steps.}
\label{tab:cost}
\small
\begin{tabular}{@{}lr@{}}
\toprule
Stage & Time per task \\
\midrule
Generating $M=40$ candidate rollouts & 11 min \\
Completion-aware truncation (40 VLM calls, $\approx$1.8M input tokens) & 3.5 min \\
Track4World flow estimation & 1.5 min \\
Robot-only simulation, scoring, and top-$N$ selection & 2 min \\
\midrule
Rectified generative replay in total & 18 min \\
Fine-tuning on the new task (1{,}000 steps) & 1.8 h \\
\bottomrule
\end{tabular}
\end{table}

\subsection{Additional Results}
\label{app:results}

\paragraph{Comparison with ReGen-style generative replay.}
\label{app:regen}
REGEN~\citep{govind2026regen} also replays generated rollouts, but initializes each rollout from a real observation of the previous task, which requires retaining those observations. To quantify what this buys, we build a ReGen-style variant of our own pipeline: each rollout is initialized from a real initial observation of the task instead of one generated from the instruction, and no consistency filtering is applied, since REGEN has no counterpart to it. REGEN ends a rollout with the value head of its Cosmos Policy backbone~\citep{cosmospolicy2026}, which LingBot-VA does not have; we therefore truncate its rollouts with the same VLM evaluator as \method, so that both variants stop at task completion. As shown in Table~\ref{tab:regen}, the ReGen-style variant reaches 68.0\% SR on LIBERO-Spatial, far below the 94.0\% of \method, indicating that retaining real observations to start from does not substitute for regenerating and rectifying rollouts from the instruction alone.

\begin{table}[t]
\centering
\caption{\textbf{Comparison with ReGen-style generative replay} on LIBERO-Spatial. The ReGen-style variant initializes each rollout from a real observation of the task and applies no consistency filtering; both variants are truncated at task completion by the same VLM evaluator.}
\label{tab:regen}
\small
\begin{tabular}{@{}lccc@{}}
\toprule
\textbf{Variant} & SR $\uparrow$ & NBT $\downarrow$ & Plast.\ $\uparrow$ \\
\midrule
ReGen-style replay & 68.0 & 19.15 & 92.2 \\
\method (ours) & \textbf{94.0} & \textbf{2.90} & \textbf{95.7} \\
\bottomrule
\end{tabular}
\end{table}

\begin{figure}[t]
\centering
\includegraphics[width=0.92\linewidth]{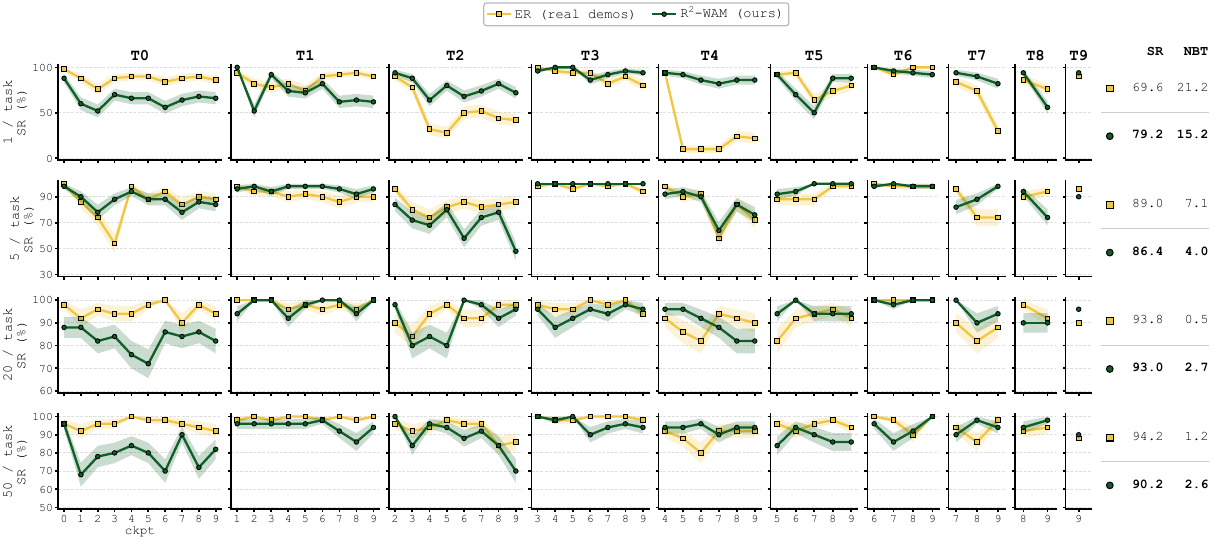}
\caption{\textbf{Task matrices when scaling the replay buffer on LIBERO-Spatial.} Rows are replay buffer sizes of 1, 5, 20, and 50 episodes and columns are the ten tasks. Each panel shows how the success rate of one task evolves as the model continues to learn the following tasks, and the last column lists the final SR and NBT of each row. ER (yellow) replays real demonstrations, and \method (green) replays generated rollouts without real data. Shading is the standard error.}
\label{fig:budget}
\end{figure}

\paragraph{Replay buffer task matrices.}
\label{app:budget}
Fig.~\ref{fig:budget} shows the per-task success curves behind Fig.~\ref{fig:budget-curves}. At a buffer size of 1, the deficit of ER comes from a few tasks (T2, T4, and T7) that collapse to 22--42\% once later tasks are learned, whereas \method degrades mildly on several tasks but never falls below 50\% on any of them, which supports the conjecture that a generated rollout is a more stable anchor than a single real demonstration. From 20 episodes onwards, ER stays above 80\% on every task, while the remaining gap of \method concentrates on a few early tasks, most notably T0, whose success keeps drifting slowly as more tasks are learned.

\subsection{Qualitative Real-Robot Rollouts}
\label{app:real-qual}
\begin{figure}[t]
\centering
\includegraphics[width=\linewidth]{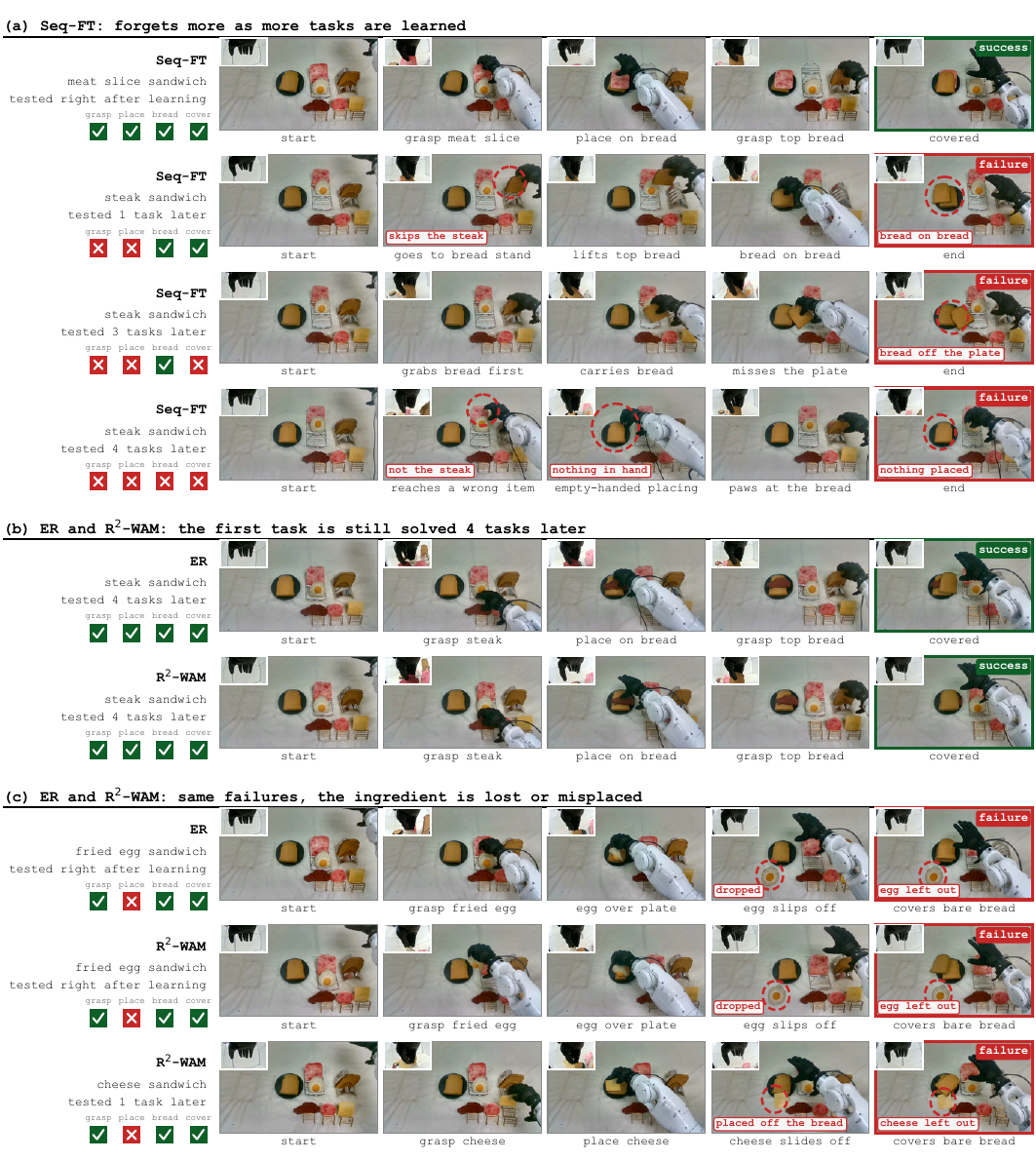}
\caption{\textbf{Qualitative real-robot rollouts.} The robot makes a sandwich with the instructed ingredient in four steps: grasp the ingredient, place it on the bread, grasp the top bread, and cover the sandwich. Each row is one trial from the head camera with the wrist view inset. Its label gives the method, the instruction, and how many further tasks were learned before testing. The four boxes mark which steps the robot completes, and red ellipses mark the error.}
\label{fig:real-cases}
\end{figure}

Fig.~\ref{fig:real-cases} shows representative trials behind Table~\ref{tab:real_robot}, from which we draw three observations. \textbf{First}, Seq-FT is highly plastic but forgets, and the forgetting worsens as more tasks are learned, as shown from top to bottom in Fig.~\ref{fig:real-cases}(a). Right after learning a task, the robot completes all four steps. After one more task is learned, it already skips the steak and completes only the two bread steps. As training continues, the bread steps degrade as well: three tasks later the robot still grasps the bread but releases it off the plate, and four tasks later it completes no step at all. \textbf{Second}, ER and \method both retain previous tasks, and four tasks later both still complete the first task with the same motion, as shown in Fig.~\ref{fig:real-cases}(b). \textbf{Finally}, both replay methods still lose some precision on previous tasks, and their failures are alike. In all three trials of Fig.~\ref{fig:real-cases}(c), the robot grasps the instructed ingredient and completes both bread steps, and fails only at placing: the fried egg slips from the hand in transit, for ER and \method alike, or the cheese is released at the edge of the bread and slides off. This loss of placement precision is consistent with the task interference discussed in Section~\ref{sec:main-results}.

\FloatBarrier
\subsection{Prompt of the Completion Evaluator}
\label{app:prompt}
The full prompt of the completion evaluator is reproduced in Fig.~\ref{fig:prompt}, instantiated for a LIBERO-Spatial task.

\newcommand{\pnote}[1]{\colorbox{black!7}{\rmfamily\color{black!60}#1}}
\newcommand{\pframe}[3]{\begin{minipage}[t]{0.235\linewidth}\centering
{\scriptsize\ttfamily #2}\\[1pt]\includegraphics[width=\linewidth]{pf_#1.pdf}\\[0pt]{\scriptsize\ttfamily #3}\end{minipage}}
\newcommand{\pframeN}[3]{\begin{minipage}[t]{0.235\linewidth}\centering
{\scriptsize\ttfamily #2}\\[1pt]\includegraphics[width=\linewidth]{pf_#1.pdf}\\[0pt]{\scriptsize\pnote{\itshape #3}}\end{minipage}}
\newcommand{\phead}[1]{\medskip\noindent\pnote{\textbf{\small #1}}\par\smallskip}
\begin{tcolorbox}[enhanced, breakable, colback=white, colframe=black!55, boxrule=0.5pt, arc=1pt,
 left=6pt, right=6pt, top=4pt, bottom=4pt, before skip=8pt, after skip=8pt]
\footnotesize\ttfamily\setlength{\parskip}{2pt}
\phead{System (system message)}
You are labelling where a robot manipulation episode should be CUT: everything after your chosen frame will be deleted from the training data. You will be shown still frames in chronological order. Each frame is preceded by a line of text giving its absolute STEP index, the GRIPPER command at that step, and the commanded end-effector VELOCITY vel=[x,y,z] (\texttt{z>0} is upward) with its speed. Gripper scale: dimension 7 of the action vector, in $[-1,+1]$; near $+1$ commands the fingers to close, near $-1$ to open; a sustained switch from $+1$ to $-1$ right after an object reaches its target is what a deliberate release looks like. The robot runs at 20 steps per second. Each image shows two views of the same moment: LEFT = external camera, RIGHT = wrist camera. FIRST work out, from the instruction alone, what the world must look like for it to be fully carried out, its GOAL STATE. THEN find where the episode reaches it. FINALLY answer with a CUT POINT inside the acceptable zone: after the motion that completes the task has fully finished, and before the arm clearly moves on to anything else. Answer with JSON only.

\phead{User (user message)}
\texttt{TASK GIVEN TO THE ROBOT: "pick up the black bowl between the plate and the ramekin and place it on the plate"}

\phead{Part 1: Annotated example}
\texttt{=== PART 1: ANNOTATED EXAMPLE} --- one complete episode from the SAME generator and SAME task, whose correct cut zone was marked by a human. Frames are sparse early and denser as the cut zone approaches, and the discarded junk after it is shown too. Each frame is labelled task still in progress, CUT ZONE (any cut inside this zone is correct: it never severs the completing motion and never keeps off-task motion), or DISCARDED (junk after the cut). Use it to calibrate what completion, the cut zone, and junk look like for this task on generated data.

\noindent\texttt{--- ANNOTATED EXAMPLE: correct cut zone = steps 92-100}\par\smallskip
\noindent\pframe{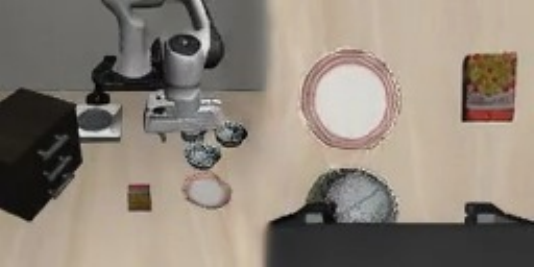}{EXAMPLE step 27\\gripper -0.99 \textbar{} speed 0.97}{task still in progress}\hfill
\pframe{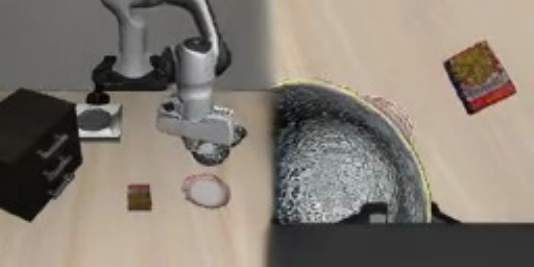}{EXAMPLE step 66\\gripper +1.00 \textbar{} speed 0.78}{task still in progress}\hfill
\pframe{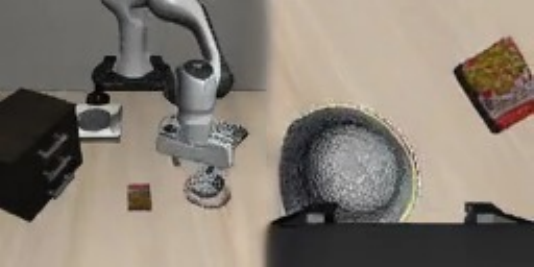}{EXAMPLE step 96\\gripper -0.98 \textbar{} speed 0.64}{*** CUT ZONE: any cut in this zone is correct ***}\hfill
\pframe{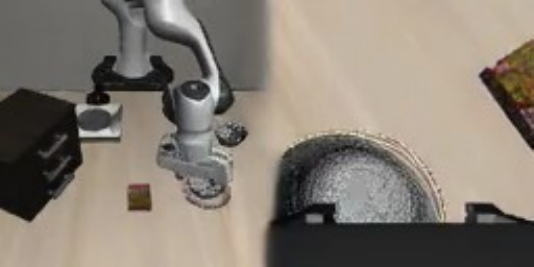}{EXAMPLE step 106\\gripper -0.98 \textbar{} speed 0.84}{DISCARDED: junk after the cut}\par\smallskip
\noindent\pnote{\scriptsize 4 of the 12 example frames are shown; the line under each frame is the label the prompt attaches to it.}

\phead{Part 2: Candidate}
\texttt{=== PART 2: CANDIDATE} --- the episode you must label, produced by a world model that always runs a fixed number of steps with no stopping criterion, so it usually keeps acting after the task is already complete: hovering, drifting, or picking the object up again. The full episode is 128 steps (6.4\,s); the task is known to be still unfinished early on, so steps 0--51 are not shown. The 39 candidate frames below are in chronological order and are of two kinds, marked in the line above each frame: \texttt{[BOUNDARY]}, a motion turning point detected from the robot's own action stream (the moment one continuous motion has just ended, a gripper release, or the final frame); and \texttt{[SAMPLE]}, a uniform time sample between boundaries. A motion segment table from the same action stream is given right before the frames. Your answer must be one of the candidate steps below.

\noindent\texttt{MOTION SEGMENTS detected from the robot's commanded end-effector velocities:}\\
\texttt{steps 48-76: mean direction [x+0.86, y+0.23, z-0.45], peak speed 1.08}\\
\texttt{steps 81-92: mean direction [x+0.29, y+0.00, z+0.96], peak speed 0.75}\\
\texttt{\ldots}\par\smallskip
\noindent\pframeN{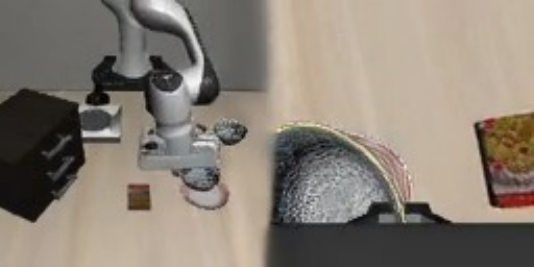}{CANDIDATE [SAMPLE] step 66\\gripper +1.00 \textbar{} speed 0.79}{carrying the bowl}\hfill
\pframeN{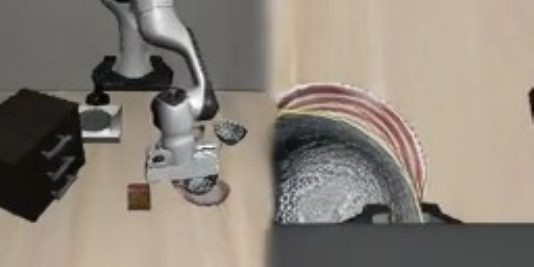}{CANDIDATE [BOUNDARY] step 78\\gripper -0.98 \textbar{} speed 0.24}{release}\hfill
\pframeN{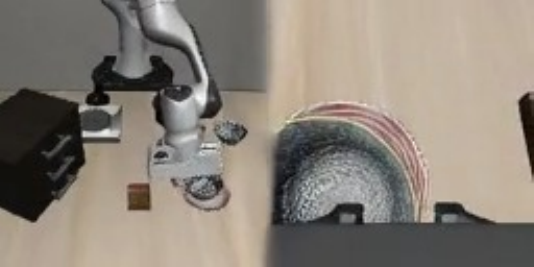}{CANDIDATE [BOUNDARY] step 82\\gripper -0.98 \textbar{} speed 0.66}{retreat}\hfill
\pframeN{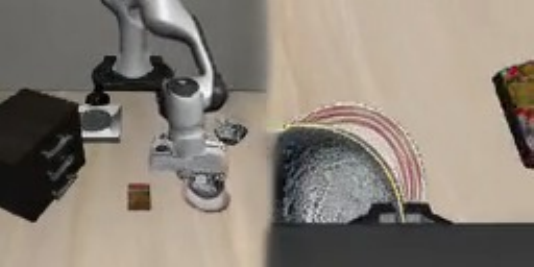}{CANDIDATE [BOUNDARY] step 126\\gripper +0.97 \textbar{} speed 0.50}{re-grasping (final frame)}\par\smallskip
\noindent\pnote{\scriptsize 4 of the 39 candidate frames are shown; the italic notes under them describe what happens and are ours.}

\phead{How to decide}
\texttt{=== HOW TO DECIDE}\par
STEP 1. Derive the goal state from the instruction before looking for an answer: list every condition that must hold at the end, in the order they must happen; the task is finished only when the last one holds. Do not assume a sub-goal involves the gripper at all.\par
STEP 2. Decide when each sub-goal is satisfied. A PLACEMENT sub-goal is satisfied only when the object rests at its target, the printed gripper values have switched from closing to opening, and the object has fully SETTLED; the instant of release is not the end. For a NON-HELD sub-goal (pushing, sliding, turning) the gripper values are irrelevant: follow the interaction motion to its TURNING POINT, the last frame before the arm switches to a different kind of motion. A sub-goal is satisfied at the LAST successful attempt, not the first try.\par
STEP 3. Pick \texttt{end\_step} from the evidence above: the annotated example, so that your episode is cut at the analogous place and keeps no frames resembling its discarded ones; the numbers, because a motion lasts exactly as long as its velocity direction persists, so never cut while that direction still holds; and the candidate types, since the correct cut is usually the \texttt{[BOUNDARY]} at or shortly after the completing motion finishes. Tie-break: when hesitating between two adjacent candidates, always take the LATER one. Self-check: if the last sub-goal is a placement and the gripper at your chosen step still commands closing, your answer cuts before the release and is wrong. If some sub-goal is never visibly satisfied in any offered frame, set \texttt{accomplished=false} and \texttt{end\_step} to the last candidate step.

\phead{Output (JSON only)}
\texttt{\{"goal\_state": "<what the scene must look like, <= 25 words>",}\\
\texttt{\ "subgoals": [\{"what": "<one condition>", "satisfied\_at\_step": <candidate step or null>\}],}\\
\texttt{\ "end\_step": <one of the candidate steps>, "accomplished": <true|false>,}\\
\texttt{\ "gripper\_open\_at\_end": <true|false>, "failed\_grasp\_attempts": <int>, "why": "<= 20 words"\}}
\end{tcolorbox}
\par\vspace{4pt}
\captionof{figure}{\textbf{The prompt of the completion evaluator.} Shown for a LIBERO-Spatial task, with four frames of the in-context example and of the candidate; some non-essential content is omitted, and shaded gray text is our annotation.}
\label{fig:prompt}

\subsection{Task Sequences}
\label{app:task-order}
The task sequence of every benchmark is listed below in learning order. Within each LIBERO suite, tasks are ordered lexicographically by their language instructions, so the order is fixed by the benchmark rather than chosen by us; the five real-robot tasks follow a fixed order that is identical for all compared methods. LIBERO tasks are indexed from T0 following the benchmark and real-robot tasks from T1.

\begin{taskbox}{LIBERO-Spatial Task Order}
\begin{enumerate}\setcounter{enumi}{-1}
 \item Pick up the black bowl between the plate and the ramekin and place it on the plate.
 \item Pick up the black bowl from table center and place it on the plate.
 \item Pick up the black bowl in the top drawer of the wooden cabinet and place it on the plate.
 \item Pick up the black bowl next to the cookie box and place it on the plate.
 \item Pick up the black bowl next to the plate and place it on the plate.
 \item Pick up the black bowl next to the ramekin and place it on the plate.
 \item Pick up the black bowl on the cookie box and place it on the plate.
 \item Pick up the black bowl on the ramekin and place it on the plate.
 \item Pick up the black bowl on the stove and place it on the plate.
 \item Pick up the black bowl on the wooden cabinet and place it on the plate.
\end{enumerate}
\end{taskbox}

\begin{taskbox}{LIBERO-Object Task Order}
\begin{enumerate}\setcounter{enumi}{-1}
 \item Pick up the alphabet soup and place it in the basket.
 \item Pick up the bbq sauce and place it in the basket.
 \item Pick up the butter and place it in the basket.
 \item Pick up the chocolate pudding and place it in the basket.
 \item Pick up the cream cheese and place it in the basket.
 \item Pick up the ketchup and place it in the basket.
 \item Pick up the milk and place it in the basket.
 \item Pick up the orange juice and place it in the basket.
 \item Pick up the salad dressing and place it in the basket.
 \item Pick up the tomato sauce and place it in the basket.
\end{enumerate}
\end{taskbox}

\begin{taskbox}{LIBERO-Goal Task Order}
\begin{enumerate}\setcounter{enumi}{-1}
 \item Open the middle drawer of the cabinet.
 \item Open the top drawer and put the bowl inside.
 \item Push the plate to the front of the stove.
 \item Put the bowl on the plate.
 \item Put the bowl on the stove.
 \item Put the bowl on top of the cabinet.
 \item Put the cream cheese in the bowl.
 \item Put the wine bottle on the rack.
 \item Put the wine bottle on top of the cabinet.
 \item Turn on the stove.
\end{enumerate}
\end{taskbox}

\begin{taskbox}{LIBERO-10 Task Order}
\begin{enumerate}\setcounter{enumi}{-1}
 \item Pick up the book and place it in the back compartment of the caddy.
 \item Put both moka pots on the stove.
 \item Put both the alphabet soup and the cream cheese box in the basket.
 \item Put both the alphabet soup and the tomato sauce in the basket.
 \item Put both the cream cheese box and the butter in the basket.
 \item Put the black bowl in the bottom drawer of the cabinet and close it.
 \item Put the white mug on the left plate and put the yellow and white mug on the right plate.
 \item Put the white mug on the plate and put the chocolate pudding to the right of the plate.
 \item Put the yellow and white mug in the microwave and close it.
 \item Turn on the stove and put the moka pot on it.
\end{enumerate}
\end{taskbox}

\begin{taskbox}{Real-Robot Task Order}
\begin{enumerate}
 \item Make a sandwich with steak.
 \item Make a sandwich with luncheon meat.
 \item Make a sandwich with cheese.
 \item Make a sandwich with a fried egg.
 \item Make a sandwich with a meat slice.
\end{enumerate}
\end{taskbox}

\end{document}